\documentclass[letterpaper, 10 pt, conference]{ieeeconf}

\IEEEoverridecommandlockouts            

\let\labelindent\relax                  

\usepackage[utf8]{inputenc}
\usepackage{graphicx} 
\usepackage{amsmath,amsfonts,amssymb,booktabs,cite,siunitx} 
\usepackage[hidelinks]{hyperref} 
\usepackage[nolist]{acronym}  
\usepackage[inline]{enumitem} 
\usepackage{caption,subcaption} 
\usepackage{flushend} 
\usepackage{pgfplots} 
\pgfplotsset{compat=1.8}

\newacro{mod}[MoD]{map of dynamics}
\newacroplural{mod}[MoDs]{maps of dynamics}
\newacro{mse}[MSE]{mean squared error}

\newcommand{\figref}[1]{\hyperref[#1]{Fig.~\ref*{#1}}}
\newcommand{\tabref}[1]{\hyperref[#1]{Tab.~\ref*{#1}}}
\newcommand{\secref}[1]{\hyperref[#1]{Sec.~\ref*{#1}}}

\newcommand{\matr}[1]{\mathbf{#1}}
\newcommand{\prob}[1]{\mathsf{P}(#1)}

\def\sota{state-of-the-art}
\def\gt{groundtruth}
\def\ie{\textit{i.e.},}
\def\eg{\textit{e.g.},}
\def\etal{\textit{et al.}}

\def\occa{s^A}
\def\occk{s^K}
\def\atct{\emph{ATC-W}}
\def\atcv{\emph{ATC-S}}
\def\fft{\matr{\hat{d}}^W}
\def\ffv{\matr{\hat{d}}^S}
\def\ffk{\matr{\hat{d}}^K}
\def\kth{\emph{KTH}}

\title{\LARGE \bf Bayesian Floor Field:\\
    Transferring people flow predictions across environments}

\author{ Francesco~Verdoja, Tomasz~Piotr~Kucner, Ville~Kyrki%
    \thanks{F.~Verdoja, T.~Kucner and V.~Kyrki are with School of Electrical
    Engineering, Aalto University, Finland. (\texttt{name.surname{@}aalto.fi})}
    }

\begin{document}

\makeatletter
\let\@oldmaketitle\@maketitle
\renewcommand{\@maketitle}{\@oldmaketitle
  \setcounter{figure}{0}
  \vspace{1em}
  \centering
  \includegraphics[width=\linewidth]{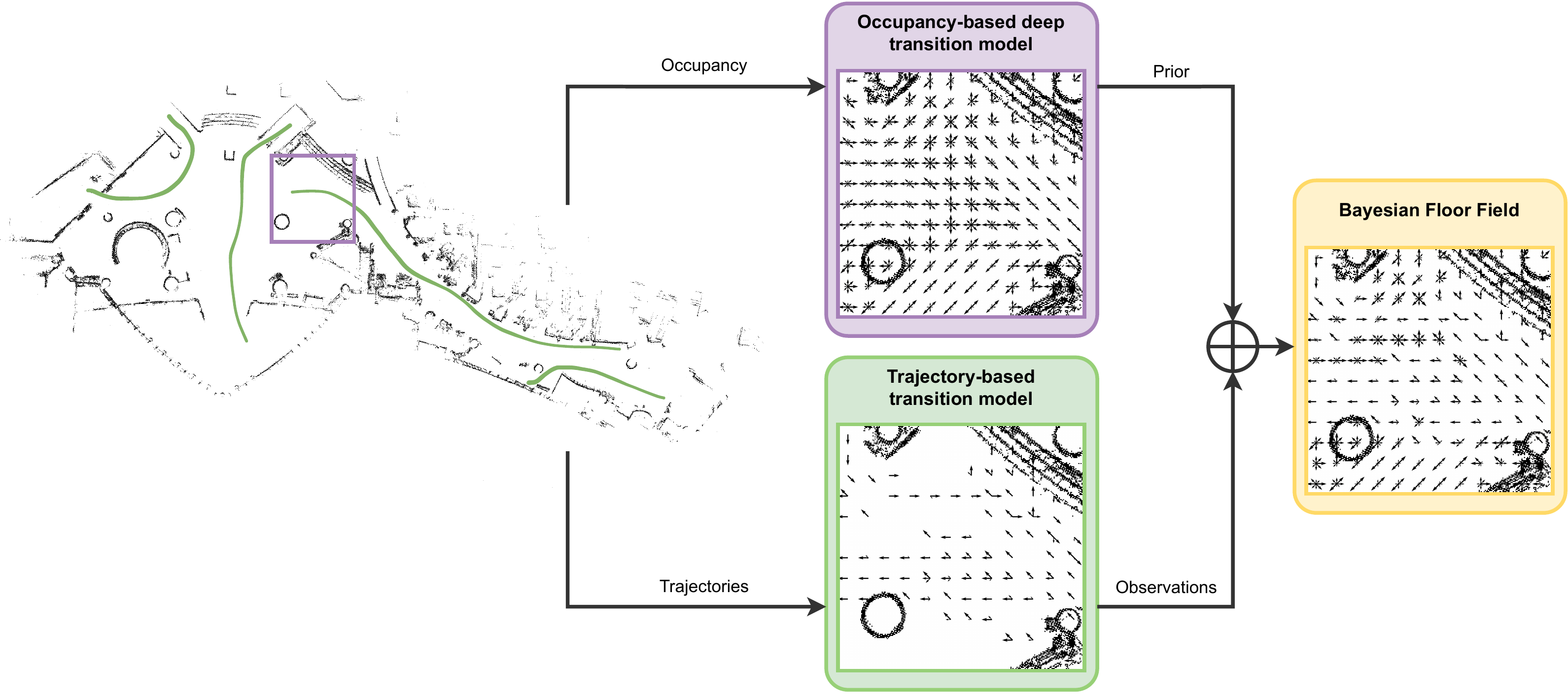}
  \captionof{figure}{\label{fig:cover}We propose Bayesian Floor Field, a method
  to build maps of human dynamics by combining information from static occupancy
  and pedestrian trajectories. Even when no trajectory data is available, our
  method can generalize to unseen environments by leveraging the relationship
  between environment geometry and human motion, reducing the need for
  time-consuming pedestrian data acquisition.}}
\makeatother

\maketitle
\thispagestyle{empty}
\pagestyle{empty}


\begin{abstract}
    Mapping people dynamics is a crucial skill for robots, because it enables
    them to coexist in human-inhabited environments. However, learning a model
    of people dynamics is a time consuming process which requires observation of
    large amount of people moving in an environment. Moreover, approaches for
    mapping dynamics are unable to transfer the learned models across
    environments: each model is only able to describe the dynamics of the
    environment it has been built in. However, the impact of architectural
    geometry on people's movement can be used to anticipate their patterns of
    dynamics, and recent work has looked into learning maps of dynamics from
    occupancy. So far however, approaches based on trajectories and those based
    on geometry have not been combined. In this work we propose a novel Bayesian
    approach to learn people dynamics able to combine knowledge about the
    environment geometry with observations from human trajectories. An
    occupancy-based deep prior is used to build an initial transition model
    without requiring any observations of pedestrian; the model is then updated
    when observations become available using Bayesian inference. We demonstrate
    the ability of our model to increase data efficiency and to generalize
    across real large-scale environments, which is unprecedented for maps of
    dynamics.
\end{abstract}

\section{Introduction}
\label{sec:intro}

In recent years, an increasing number of robots are being deployed in shared
environments, where humans and autonomous agents coexist and collaborate. To
assure the safety and efficiency of robots' operation in such environments, it
is necessary to enable robots to understand and adhere to shared norms,
including the implicit traffic rules governing the motion of
pedestrians~\cite{Kruse2013}.

Over the past years, we have observed the development of different methodologies
for modelling pedestrian motion, one of which is
\acp{mod}~\cite{kucner_survey_2023}. \acp{mod} capture the common motion
patterns followed by uncontrolled agents (\ie{} humans, human-driven vehicles)
in the environment and enable robots to anticipate typical behaviors throughout
the environment. Unfortunately, the process of building \acp{mod} is very time-
and resource-consuming: reliable \acp{mod} are built through measuring repeating
motion patterns executed by uncontrolled agents in the given environment. As a
consequence, the deployment of a successful robotic system using \acp{mod}
requires a substantial amount of time necessary to collect enough relevant
data~\cite{Kucner2020,kucner_survey_2023}. Moreover, a \ac{mod} is only able to
describe and predict pedestrian motion in the same environment it has been built
in. The inability to transfer between environments is a crucial limitation of
\acp{mod}, especially considering that pedestrian traffic rules share
commonalities across environments, \eg{} people move similarly through a
corridor, or around a door.

To address this limitation, recently we have seen the development of methods
that leverage the correlation between the shape of the environment and the
behavior of humans therein, to predict the possible motion patterns in it. That
said, the existing efforts have been narrow in scope, primarily focusing either
on the use of synthetic trajectory data~\cite{Lai2019, Zhi2019}, or being
limited to small environments~\cite{doellinger_predicting_2018,
doellinger_environment-aware_2019}. Moreover, how to combine these
occupancy-based methods with observations of human movement has not yet been
studied.

In this work we propose a novel Bayesian \ac{mod} approach using people dynamics
learned from environment occupancy as prior, and updating the model with
measured human trajectories. As opposed to previous efforts, we train our
occupancy-based prior on real human data in a real large-scale environment and
evaluate the capability of the learned prior to transfer across environments. An
illustration of the approach is given in \figref{fig:cover}.

In particular, the contributions of this work are:
\begin{enumerate}
    \item A novel approach for training occupancy-based deep transition
          probability models from real data;
    \item A novel method for building \acp{mod} by combining knowledge about the
          environment geometry with observations of people motion;
    \item A study over the ability of the proposed occupancy-based approach to
          model real unseen pedestrian motion both in the same environment it
          was trained on, as well as in a completely different large-scale
          environment never seen during training;
    \item Experimental evidence that knowledge about the environment occupancy
          can reduce the amount of trajectory data required to build \acp{mod},
          by comparing the proposed method against a traditional \ac{mod}
          approach;
    \item Source code for both our full Bayesian \ac{mod} method as well as the
          first openly available source code and trained models to learn
          \acp{mod} from occupancy\footnote{Source code, trained models, and
          data used in this paper can be found at
          \href{https://github.com/aalto-intelligent-robotics/directionalflow}
          {github.com/aalto-intelligent-robotics/directionalflow}.}.
\end{enumerate}

\section{Related work}
\label{sec:related}

The observation that people tend to follow spatial or spatio-temporal patterns
enabled the development of \acp{mod}. \acp{mod} are a special case of semantic
maps, where information about motion patterns is retained as a feature of the
environment. The existing representations can be split into three
groups~\cite{kucner_survey_2023}:
\begin{enumerate*}[label=(\arabic*)]
    \item \emph{trajectory maps}, where the information about the motion
          patterns is retained as a mixture model over the trajectory
          space~\cite{Bennewitz2002, Morris2008};
    \item \emph{directional maps}, where dynamics are represented as a set of
          local mixture models over velocity space~\cite{Senanayake2017,
          Kucner2020, Molina2018, shi_learning_2023}; and
    \item \emph{configuration changes maps}, in contrast to previously mentioned
          maps, this type of representations does not retain information about
          motion but instead, it presents the pattern of changes caused by
          semi-static objects~\cite{Krajnik2016, Saarinen2012}.
\end{enumerate*}

In this work, we are especially focusing on directional maps, which are well
suited to represent local dynamic patterns caused by directly observed moving
agents while being robust against partial or noisy observations. Furthermore,
this type of \acp{mod} consists of a large spectrum of representations of
varying levels of expressiveness and complexity. Including fairly simple models
such as floor fields~\cite{Burstede2001} as well as more complex multimodal,
continuous representations~\cite{Kucner2017}.

It is also important to emphasize that, dynamics do not exist in a vacuum but
are affected by environmental conditions, primarily by the environment's
geometry. This idea was initially presented by Helbing and
Molnar~\cite{Helbing1995} and later successfully utilized to solve the problem
of motion prediction~\cite{Rudenko2020}.

Even though the idea of utilizing metric information to inform dynamics has
substantially impacted the motion prediction community it has not yet received
adequate attention in the field of \acp{mod}. One of the more impactful attempts
in this direction is the work by Zhi \etal{}~\cite{Zhi2019}. In that work, the
authors utilize artificially generated trajectories to train a deep neural
network to predict possible behaviors in new unobserved environments.

At this same time, Doelinger
\etal{}~\cite{doellinger_predicting_2018,doellinger_environment-aware_2019}
proposed a method to predict not the motion itself but the levels of possible
activity in given environments, based on surrounding occupancy. Both works by
Zhi \etal{}~\cite{Zhi2019} and Doelinger
\etal{}~\cite{doellinger_predicting_2018, doellinger_environment-aware_2019}
present important steps towards the prediction of motion patterns given the
environment geometry.

However, the aforementioned contributions are application specific and narrow in
scope, using either only synthetic data, or by being limited to only small
environments. In this work, we propose a step change with respect to the
presented \sota{} by presenting a way to leverage real human trajectories in
large-scale environments and open new possibilities for predicting not only
motion patterns but other environment-dependent semantics. Moreover, we present
the first approach to combine occupancy-based and trajectory-based \acp{mod}.

\section{Method}
\label{sec:method}

\begin{figure*}
    \includegraphics[width=\linewidth]{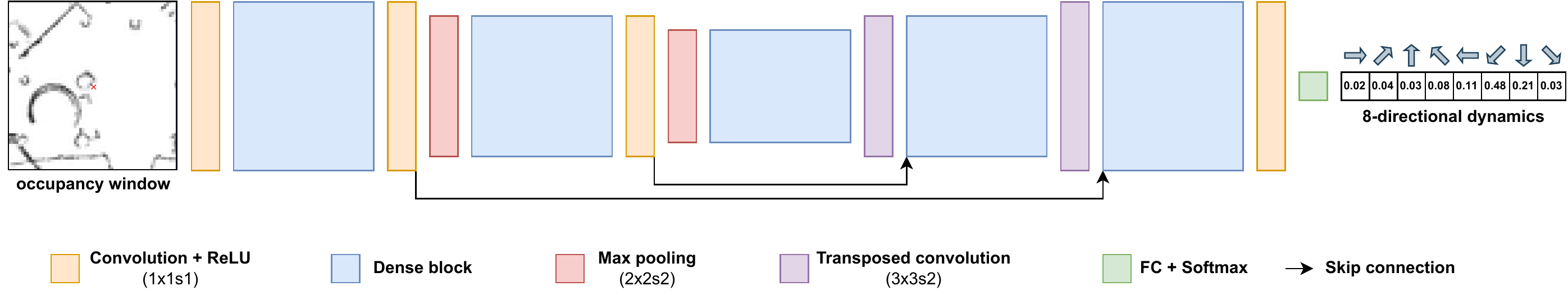}
    \caption{\label{fig:net}The network used in this work to map from
        \numproduct[mode=text]{64x64} occupancy windows to transition
        probabilities at the center of the window (marked in red).}
\end{figure*}

Let us represent the environment $\matr{M}$ the robot is operating in as a
collection of cells $c \in \matr{M}$. For each cell $c$, we assume to know an
occupancy probability $s(c) \in [0,1]$ describing the likelihood of that portion
of the environment to be occupied by static objects. We refer to $s$ as the
\emph{static occupancy map} of the environment built
following~\cite{moravec_high_1985}.

To model people movement in the environment, for each cell $c$, we want to
determine the likelihood that a person in $c$ will head in a particular
direction $\delta \in [0,2\pi)\,\unit{\radian}$. Formally, we define the
transition model for cell $c$ as a categorical distribution
$\operatorname{Cat}(k, \matr{d}_c)$ over $k$ discrete directions equally
dividing the range $[0, 2\pi)\,\unit{\radian}$, \ie{}
\begin{equation}\label{eq:mod}
    \prob{\delta \mid \matr{d}_c} =
    \sum_{i=1}^k d_{ic}{\mathbf{1}_i(\delta)}\enspace,
\end{equation}
where $\matr{d}_c = (d_{1c},\dots, d_{kc} \mid \sum_{i=1}^k d_{ic} = 1)$,
$d_{ic}$ represents the probability of moving toward direction $i$ from $c$, and
the indicator function $\mathbf{1}_i(\delta) = 1$ iff $\frac{2(i-1)}{k}\pi \leq
\delta < \frac{2i}{k}\pi$, $0$ otherwise. We use $\mathbf{d}_c$ as shorthand for
$\mathbf{d}(c)$. We refer to the complete model $\matr{d}$ as the
\emph{\acl{mod}} (or \emph{people flow map}) of the environment.

Previous works building \acp{mod} as grid-based categorical distributions infer
the distribution parameter $\matr{d}_c$ for each cell $c$ from trajectory data
in a frequentistic fashion \cite{Burstede2001, Kucner2020}. Instead, in this
work we treat it as a Bayesian inference problem.

The main assumptions in this work are that:
\begin{enumerate*}[label=(\roman*)]
    \item an environment's occupancy around a certain location (\ie{} that
    location's neighborhood) influences how people move from it; and
    \item neighborhoods having similar occupancy, even from different
    environments, influence people movement similarly.
\end{enumerate*}
Under these assumptions, we treat $\matr{d}_c$ as a random variable whose
posterior is inferred by incorporating information about the environment
occupancy around $c$, used as prior, and observations coming from trajectory
data. Formally, according to assumption (i), given a certain neighborhood
$\matr{N}_{c_r} \subset \matr{M}$ around a reference cell $c_r$, our first
target is to learn the prior $\matr{\bar{d}}_{c_r} = \prob{\matr{d}_{c_r} \mid
\matr{w}_{c_r}}$, where $\matr{w}_{c_r} = \{s(c) \mid c \in \matr{N}_{c_r}\}$,
\ie{} a window over the occupancy map describing the geometry of the environment
around $c_r$. We also propose that, according to assumption (ii), given a
different environment $\matr{M}'$, and a reference cell $c'_r \in \matr{M}'$,
$\matr{w}_{c'_r} \approx \matr{w}_{c_r} \implies \matr{\bar{d}}_{c'_r} \approx
\matr{\bar{d}}_{c_r}$. 

In \secref{sec:post} we will present how to obtain the posterior for
$\matr{d}_c$, while in \secref{sec:net} we will propose a method to approximate
the prior $\matr{\bar{d}}_{c}$ through deep learning.

\subsection{Posterior inference using conjugate prior}
\label{sec:post}

Given the prior $\matr{\bar{d}}_{c}$ and a set of observations $\mathbb{D}_c =
\{\delta_1,...\delta_N\} \sim \operatorname{Cat}(k, \matr{d}_c)$, \eg{} obtained
from trajectory data of people moving in the environment, we can infer the
posterior using the Dirichlet distribution, conjugate prior for the categorical
distribution \cite{minka_bayesian_2003}.

Let $\alpha > 0$ be a concentration hyperparameter, indicating our trust in the
prior, and $\matr{q}_c = (q_{1c}, \dots, q_{kc})$ represent the number of
occurrences of direction $i$ in $\mathbb{D}_c$ such that $q_{ic} = \sum_{j=1}^N
\mathbf{1}_i(\delta_j)$. Then the posterior $\prob{\matr{d}_{c} \mid \alpha,
\matr{\bar{d}}_{c}, \mathbb{D}_c} \sim \operatorname{Dir}(k, \matr{q}_c + \alpha
\cdot \matr{\bar{d}}_c)$, which allows us to calculate the expected value for
each directional probability $d_{ic}$ as:
\begin{equation}\label{eq:postexp}
    \operatorname{E}[d_{ic} \mid \alpha, \matr{\bar{d}}_{c}, \mathbb{D}_c] = 
    \frac{q_{ic} + \alpha \cdot \bar{d}_{ic}}{N + \alpha}\enspace.
\end{equation}

The set of all $d_{ic}~\forall i \in [1,k]; c \in \matr{M}$ defines the complete
posterior $\matr{d}$. Whenever new observations become available,
\eqref{eq:postexp} is also used to updated the belief over the posterior.

\subsection{Parametric approximation of the prior}
\label{sec:net}

We learn $g_\theta \approx \prob{\matr{d} \mid \matr{w}}$, \ie{} a parametric
approximation of the prior defined by the parametrization $\theta$, that we
model as a FC-DenseNet architecture~\cite{jegou_one_2017} following previous
literature~\cite{doellinger_predicting_2018, doellinger_environment-aware_2019}.
The structure of the network is shown in \figref{fig:net}. The network takes as
input a \numproduct[mode=text]{64x64} window over an occupancy grid map,
processes it over several densely connected blocks of convolutional layers and
max-pooling layers, before up-sampling it through transposed convolutions and
outputting the $k$-dimensional transition probability distribution
$\matr{\bar{d}}_{c_r}$, with $c_r$ being the center pixel of the input window.
In this study we use $k=8$ in order to model the probability of moving in the
direction of each of the eight neighboring cells to $c_r$. Please refer
to~\cite{jegou_one_2017} for the exact composition of the dense blocks.

One thing to note is that most occupancy grid maps are built at very high
resolution (usually \qtyrange[mode=text]{0.05}{0.1}{\metre}), but for \acp{mod}
modelling human traffic, those resolutions are too dense, and they are usually
constructed at around \qtyrange[mode=text]{0.4}{1}{\metre} per
cell~\cite{Kucner2020}. Therefore, to be able to build models at arbitrary
output grid resolutions, we scale the grid resolution for the input of the
network, by interpolating from the original grid resolution of the occupancy
map. This means that, for example, if we want the network to learn a people flow
map at a \qty[mode=text]{0.4}{\metre\per cell} resolution, the
\numproduct[mode=text]{64x64} input window covers an area of
\qty[mode=text]{25.6}{\square\metre}.

\section{Implementation}
\label{sec:implement}

\subsection{Training and augmentations}
\label{sec:train}

We train $g_\theta$ in a supervised fashion by using a dataset of pairs
$(\matr{w}_c, \matr{\hat{d}}_c)$ of occupancy windows $\matr{w}_c$ with their
corresponding \gt{} transitions $\matr{\hat{d}}_c$ (more details in
\secref{sec:data}). As loss we use \ac{mse} between the predicted transition
probabilities and the \gt{}. We train for 120 epochs, using Adam as optimizer
with a fixed learning rate of \num[mode=text]{0.001}.

To extend the amount of available data, we augment each input-output pair
randomly by vertical and/or horizontal flipping followed by a random rotation of
either \num{0}, $\frac{1}{2}\pi$, $\pi$, or $\frac{3}{2}\pi\,\unit{\radian}$,
with equal probability. When we perform these augmentations, the \gt{}
transition probabilities are transformed accordingly to still match the
transformed input window.

\subsection{Datasets}
\label{sec:data}

The dataset we use for training our model is the \emph{ATC Dataset}, containing
real pedestrian data from the ATC mall (The Asia and Pacific Trade Center,
Osaka, Japan, first described by~\cite{Brvsvcic2013}). This dataset was
collected with a system consisting of multiple 3D range sensors, covering an
area of about \qty[mode=text]{900}{\square\metre}. The data was collected
between October 24, 2012 and November 29, 2013, every week on Wednesday and
Sunday between 9:40 and 20:20, which gives a total of 92 days of observation.
For each day a large number of human trajectories is recorded. An occupancy grid
map of the environment $\occa$ at \qty[mode=text]{0.05}{\metre\per pixel}
resolution is also available with the dataset.

From the ATC dataset, we picked Wednesday November 14, 2012 for training and
Saturday November 18, 2012 for testing, which we will refer to as \atct{} and
\atcv{} respectively. For each day, we built a \gt{} \ac{mod} using the floor
field algorithm~\cite{Burstede2001} which constructs a per-cell 8-directional
transition model by accumulation directly from trajectory data. We refer to
these models as $\fft$ and $\ffv$ respectively. For training then, we will use a
dataset composed of 1479 pairs of cells $(\matr{w}_c, \fft_c)$, while a dataset
of 1360 pairs $(\matr{w}_c, \ffv_c)$ will be used for validation. In both,
$\matr{w}_c$ is a window around cell $c$ extracted from $\occa$.

As dataset representing an unseen environment, we use the \emph{KTH Track
Dataset}~\cite{dondrup_real-time_2015}, which we will refer to as \kth{}. Data
from this dataset is never seen during training and is only used for evaluation.
In this dataset, 6251 human trajectory data were collected by an RGB-D camera
mounted on a Scitos G5 robot navigating through University of Birmingham
library. An occupancy grid map of the environment $\occk$ at
\qty[mode=text]{0.05}{\metre\per pixel} resolution is also available with the
dataset. Similarly to ATC, for this dataset we learn a gold standard floor field
model $\ffk$ using all trajectories available for this dataset.
\section{Experiments}
\label{sec:exp}

In order to evaluate the proposed method, we want to assess:

\begin{enumerate}
    \item how informative is the learned prior when modelling human motion
          within the same environment, but on a different day (\atcv{});
    \item how well does the Bayesian \acl{mod} approach transfer to a different
    environment (\kth{});
    \item how does the network input resolution affect generalization
    performance.
\end{enumerate}

In these experiments, we want to compare the performance of the proposed
transition model against the gold standard model, \ie{} the floor field model
built only using trajectory data, as well as a Bayesian model using an
uninformed uniform prior. 

As a metric, we will compute the likelihood for a trajectory dataset to be
predicted by each model. Formally, each trajectory dataset $\mathbb{D}$ is a
sequence of observations defined by their $xy$-coordinates and a motion angle
$\delta \in [0,2\pi)$, \ie{} $\mathbb{D} = [(x_1, y_1, \delta_1), \dots, (x_N,
y_N, \delta_N)]$ representing the position a person was standing in (in world
coordinates) and the direction they were moving towards. Technically, each
observation also has a timestamp and the person id the observation belongs to,
but we do not use either in this evaluation. Then, given a transition model
$\matr{d}$, the average likelihood for a dataset $\mathbb{D}$ is computed as
\begin{equation}\label{eq:likelihood}
    \mathcal{L}(\mathbb{D} \mid \matr{d}) =
    \frac{1}{N}\sum^{N}_{j=1}\prob{\delta_j \mid \matr{d}_{c_j}}\enspace,
\end{equation}
where $\matr{d}_{c_j}$ refers to the transition model for the grid cell ${c_j}$
containing the coordinates $(x_j, y_j)$, and $\prob{\delta_j \mid
\matr{d}_{c_j}}$ is computed following \eqref{eq:mod}.

As datasets for our evaluation we use all observations from \atcv{} and \kth{},
namely $\mathbb{D}^S$ and $\mathbb{D}^K$ respectively. $\mathbb{D}^S$ contains
\num[mode=text]{51844} trajectories, amounting to \num[mode=text]{8533469}
observations in total, while $\mathbb{D}^K$ contains \num[mode=text]{6251}
trajectories, amounting to \num[mode=text]{421111} observations in total.

As first step, we train the network-based prior used by our method on \atct{}.
To measure the impact of window resolution on the generalization performance of
the prior, we are testing three different network generating the prior, working
with an input resolution of \qtylist[mode=text]{0.4;0.8;1}{\metre\per cell}
respectively. All other training parameters are the same as presented in
\secref{sec:train}. We will refer to these three priors as
$\matr{\bar{d}}^W_{0.4}$, $\matr{\bar{d}}^W_{0.8}$, and $\matr{\bar{d}}^W_{1.0}$
respectively. Moreover, we will refer to the uniform prior as
$\matr{\bar{d}}_\mathcal{U}$. For all methods utilizing a prior, we use $\alpha
= 5$.

We want to measure the performance of each model as a function of the number of
observations used to build it. Our expectation is that our method, by relying on
the prior, will improve the performance especially under low numbers of
observations; as the number of available observation grows, the benefits
provided by the prior will diminish and the floor field model will be able to
capture the dataset equally well. In order to validate this hypothesis, we split
each dataset in chunks of \num[mode=text]{2000} observations, and then evaluate
each method after growing the dataset one chunks at a time. We will use the
notation $\mathbb{D}[n]$ to refer to the subset composed of the first $n$
observations of dataset $\mathbb{D}$. We also use the notation
BFF($\matr{\bar{d}}, \mathbb{D}[n]$) to refer to the posterior of our Bayesian
floor field method using prior $\matr{\bar{d}}$ and the $\mathbb{D}[n]$
observation dataset, and FF($\mathbb{D}[n]$) to refer to the floor field model
using the $\mathbb{D}[n]$ observation dataset.

\begin{figure*}
    \centering
    \begin{subfigure}{0.49\textwidth}
\begin{tikzpicture}

\definecolor{crimson2143940}{RGB}{214,39,40}
\definecolor{darkorange25512714}{RGB}{255,127,14}
\definecolor{darkslategray38}{RGB}{38,38,38}
\definecolor{darkslategray88}{RGB}{88,88,88}
\definecolor{forestgreen4416044}{RGB}{44,160,44}
\definecolor{lavender234234242}{RGB}{234,234,242}
\definecolor{mediumpurple148103189}{RGB}{148,103,189}
\definecolor{steelblue31119180}{RGB}{31,119,180}

\pgfplotsset{interval/.style 2 args={
  x filter/.code={
      \ifnum\coordindex<#1\def\pgfmathresult{}\fi
      \ifnum\coordindex>#2\def\pgfmathresult{}\fi
  }
}}

\begin{axis}[
axis background/.style={fill=lavender234234242},
axis line style={white},
legend cell align={left},
legend style={
  fill opacity=0.8,
  draw opacity=1,
  text opacity=1,
  at={(0.97,0.03)},
  anchor=south east,
  draw=none,
  fill=lavender234234242,
  domain=0:140000
},
tick align=outside,
tick pos=left,
x grid style={white},
xlabel=\textcolor{darkslategray38}{number of observations $n$},
xmajorgrids,
xmin=-9900, xmax=147900,
xtick distance=20000,
xtick style={color=darkslategray38},
y grid style={white},
ylabel=\textcolor{darkslategray38}{$\mathcal{L}(\mathbb{D}^S \mid \matr{d})$},
ymajorgrids,
ymin=0.18, ymax=0.242,
ytick distance=0.01,
ytick style={color=darkslategray38}
]
\addplot [line width=1pt, interval={0}{70}, steelblue31119180]
table {figures/curves/atc_s20.txt};
\addlegendentry{$\matr{d} =$ BFF($\matr{\bar{d}}^W_{1.0}, \mathbb{D}^S[n]$)}

\addplot [line width=1pt, interval={0}{70}, darkorange25512714]
table {figures/curves/atc_s16.txt};
\addlegendentry{$\matr{d} =$ BFF($\matr{\bar{d}}^W_{0.8}, \mathbb{D}^S[n]$)}

\addplot [line width=1pt, interval={0}{70}, forestgreen4416044]
table {figures/curves/atc_s8.txt};
\addlegendentry{$\matr{d} =$ BFF($\matr{\bar{d}}^W_{0.4}, \mathbb{D}^S[n]$)}

\addplot [line width=1pt, interval={0}{70}, crimson2143940]
table {figures/curves/atc_uni.txt};
\addlegendentry
{$\matr{d} =$ BFF($\matr{\bar{d}}_{\mathcal{U}}, \mathbb{D}^S[n]$)}

\addplot [line width=1pt, interval={0}{70}, mediumpurple148103189]
table {figures/curves/atc_trad.txt};
\addlegendentry{$\matr{d} =$ FF($\mathbb{D}^S[n]$)}

\addplot [line width=0.2pt, darkslategray88, dashed, samples=2]
{0.238646740563522}; 
\addlegendentry{Upper-bound}
\end{axis}

\end{tikzpicture}
        \caption{\atcv{}}\label{fig:atc-curves}
    \end{subfigure}
    \begin{subfigure}{0.49\textwidth}
\begin{tikzpicture}

\definecolor{crimson2143940}{RGB}{214,39,40}
\definecolor{darkorange25512714}{RGB}{255,127,14}
\definecolor{darkslategray38}{RGB}{38,38,38}
\definecolor{darkslategray88}{RGB}{88,88,88}
\definecolor{forestgreen4416044}{RGB}{44,160,44}
\definecolor{lavender234234242}{RGB}{234,234,242}
\definecolor{mediumpurple148103189}{RGB}{148,103,189}
\definecolor{steelblue31119180}{RGB}{31,119,180}

\pgfplotsset{interval/.style 2 args={
  x filter/.code={
      \ifnum\coordindex<#1\def\pgfmathresult{}\fi
      \ifnum\coordindex>#2\def\pgfmathresult{}\fi
  }
}}

\begin{axis}[
axis background/.style={fill=lavender234234242},
axis line style={white},
legend cell align={left},
legend style={
  fill opacity=0.8,
  draw opacity=1,
  text opacity=1,
  at={(0.97,0.03)},
  anchor=south east,
  draw=none,
  fill=lavender234234242,
  domain=0:140000
},
tick align=outside,
tick pos=left,
x grid style={white},
xlabel=\textcolor{darkslategray38}{number of observations $n$},
xmajorgrids,
xmin=-10000, xmax=150000,
xtick distance=20000,
xtick style={color=darkslategray38},
y grid style={white},
ylabel=\textcolor{darkslategray38}{$\mathcal{L}(\mathbb{D}^K \mid \matr{d})$},
ymajorgrids,
ymin=0.135, ymax=0.197,
ytick distance=0.01,
ytick style={color=darkslategray38}
]
\addplot [line width=1pt, interval={0}{70}, steelblue31119180]
table {figures/curves/kth_s20.txt};
\addlegendentry{$\matr{d} =$ BFF($\matr{\bar{d}}^W_{1.0}, \mathbb{D}^K[n]$)}

\addplot [line width=1pt, interval={0}{70}, darkorange25512714]
table {figures/curves/kth_s16.txt};
\addlegendentry{$\matr{d} =$ BFF($\matr{\bar{d}}^W_{0.8}, \mathbb{D}^K[n]$)}

\addplot [line width=1pt, interval={0}{70}, forestgreen4416044]
table {figures/curves/kth_s8.txt};
\addlegendentry{$\matr{d} =$ BFF($\matr{\bar{d}}^W_{0.4}, \mathbb{D}^K[n]$)}

\addplot [line width=1pt, interval={0}{70}, crimson2143940]
table {figures/curves/kth_uni.txt};
\addlegendentry
{$\matr{d} =$ BFF($\matr{\bar{d}}_{\mathcal{U}}, \mathbb{D}^K[n]$)}

\addplot [line width=1pt, interval={0}{70}, mediumpurple148103189]
table {figures/curves/kth_trad.txt};
\addlegendentry{$\matr{d} =$ FF($\mathbb{D}^K[n]$)}

\addplot [line width=0.2pt, darkslategray88, dashed, samples=2]
{0.193364578475351}; 
\addlegendentry{Upper-bound}
\end{axis}

\end{tikzpicture}
        \caption{\kth{}}\label{fig:kth-curves}
    \end{subfigure}
    \caption{Performance improvements by the proposed Bayesian Floor Field (BFF)
    over \sota{} Floor Field (FF), measured in function of the amount of
    observations available. For each graph, the dotted line represents the
    likelihood upper-bound (best viewed in color).}\label{fig:curves}
\end{figure*}

\figref{fig:curves} shows the performance of each method in each testing
environment as a function of $n$, \ie{} the amount of observations available. To
improve readability of the plots, we crop the visualization at $n = 140000$;
after that point each method has had enough data to converge and performs
approximately the same. \num[mode=text]{140000} observations correspond to
approximately \num[mode=text]{850} trajectories on \atcv{} and
\num[mode=text]{2078} trajectories in \kth{}.

One thing to note about the metric used, the likelihood $\mathcal{L}$ has an
dataset-specific upper-bound, given by the intrinsic ambiguity of behaviors in
each dataset (it is impossible to predict with absolute certainty the direction
a person in a certain location will move towards). These upper-bounds are
reached when using all available trajectories to build the transition model and
then measuring the likelihood of those same trajectories given those complete
models. These upper-bounds are $\mathcal{L}(\mathbb{D}^S \mid \ffv) = 0.2386$
for \atcv{} and $\mathcal{L}(\mathbb{D}^K \mid \ffk) = 0.1933$ for \kth{}
respectively, and are shown as a dotted line in \figref{fig:curves}.

When considering \atcv{} (\figref{fig:atc-curves}), \ie{} the same environment
the network was trained on, but using trajectories from a different day, we
immediately can notice that adding an uninformed prior, \ie{}
$\matr{\bar{d}}_\mathcal{U}$, considerably hinders the performance. However, we
can see that our learned priors perform much better. In this scenario,
$\matr{\bar{d}}^W_{0.8}$ and $\matr{\bar{d}}^W_{1.0}$ only marginally improve
the performance of the floor field model, while $\matr{\bar{d}}^W_{0.4}$
improves the performance up to $n = 100000$, corresponding to the first 608
trajectories. 

When looking at the generalization capabilities of our proposed method on the
unseen \kth{} environment (\figref{fig:kth-curves}), we can make a couple of
observations. Generalization seems highly impacted by grid resolution, with
$\matr{\bar{d}}^W_{1.0}$ and $\matr{\bar{d}}^W_{0.8}$ being actually able to
generalize and improve the performance over floor field model up to $n = 20000$
(297 trajectories), while $\matr{\bar{d}}^W_{0.4}$ cannot. 

\begin{table}
    \centering
    \caption{\label{tab:results}Average likelihood and percentage
    of the range between lower and upper bounds at $n=0$ when using different priors.}
    \small
    \begin{tabular}{lcccc}
        \toprule
                                     & \multicolumn{2}{c}{\atcv{}}    & \multicolumn{2}{c}{\kth{}}     \\
        Prior                        & $\mathcal{L}$  & \%            & $\mathcal{L}$  & \%            \\
        \midrule
        $\matr{\bar{d}}_\mathcal{U}$ &         0.125  &          0.0  &         0.125  &          0.0  \\
        $\matr{\bar{d}}^W_{0.4}$     & \textbf{0.206} & \textbf{73.0} &         0.126  &          1.5  \\
        $\matr{\bar{d}}^W_{0.8}$     &         0.184  &         53.2  &         0.138  &         19.1  \\
        $\matr{\bar{d}}^W_{1.0}$     &         0.189  &         57.7  & \textbf{0.151} & \textbf{38.2} \\
        \midrule
        Upper-bound                  &         0.236  &        100.0  &         0.193  &        100.0  \\
        \bottomrule
    \end{tabular}
\end{table}

Another interesting observation comes when looking at $n = 0$ in
\figref{fig:kth-curves}, \ie{} when only the prior is used in a completely
unseen environment with no data from that domain. \tabref{tab:results} presents
the results for $n=0$ more explicitly. It can be seen that all occupancy-based
priors are performing better than the uninformed one, with
$\matr{\bar{d}}^W_{1.0}$ showing the highest likelihood, \ie{} $\mathcal{L}
=0.151$, which is around 38\% better, when considering the range up to the
upper-bound for \kth{}. This is a remarkable result that demonstrates at the
fact that knowledge about the relationship between environment occupancy and
people flow transfer across environments. 

When looking at the differences in performance between the different priors
across both tests we can conclude that $\matr{\bar{d}}^W_{0.4}$ is able to
overfit the training environment and therefore its generalization capabilities
are hindered, while the other two models demonstrate lower performance on the
same environment they were trained on, but improved generalization capabilities
on the novel unseen environment, \ie{} \kth{}.

\begin{figure*}
    \centering
    \setlength{\fboxsep}{0pt}
    \setlength{\fboxrule}{0.2pt}
    \begin{subfigure}[t]{0.24\textwidth}
        \fbox{\includegraphics[width=\linewidth]{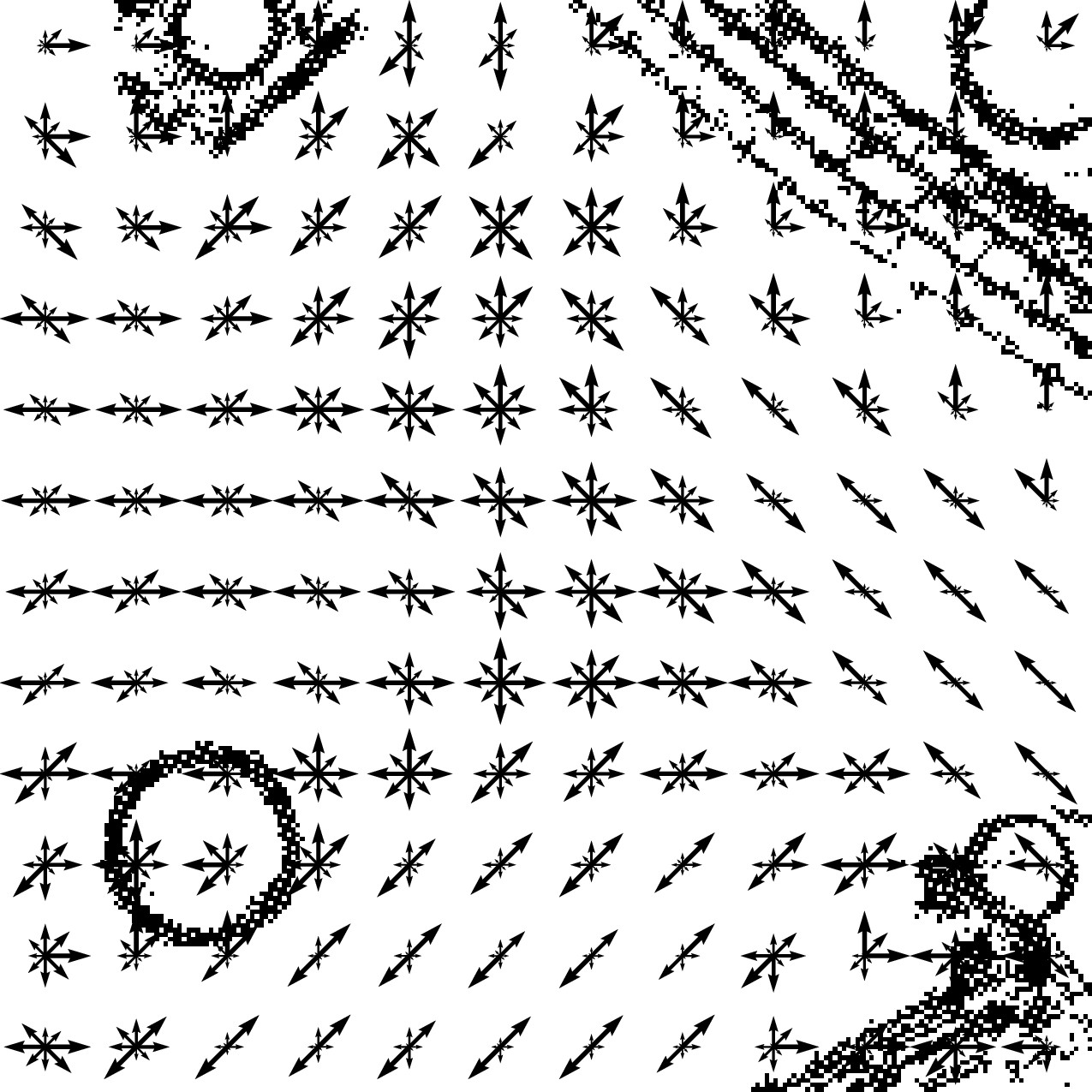}}
        \caption{ATC: Prior only ($\matr{\bar{d}}^W_{1.0}$)}%
        \label{fig:quivers-atc-prior}
    \end{subfigure}
    \begin{subfigure}[t]{0.24\textwidth}
        \fbox{\includegraphics[width=\linewidth]{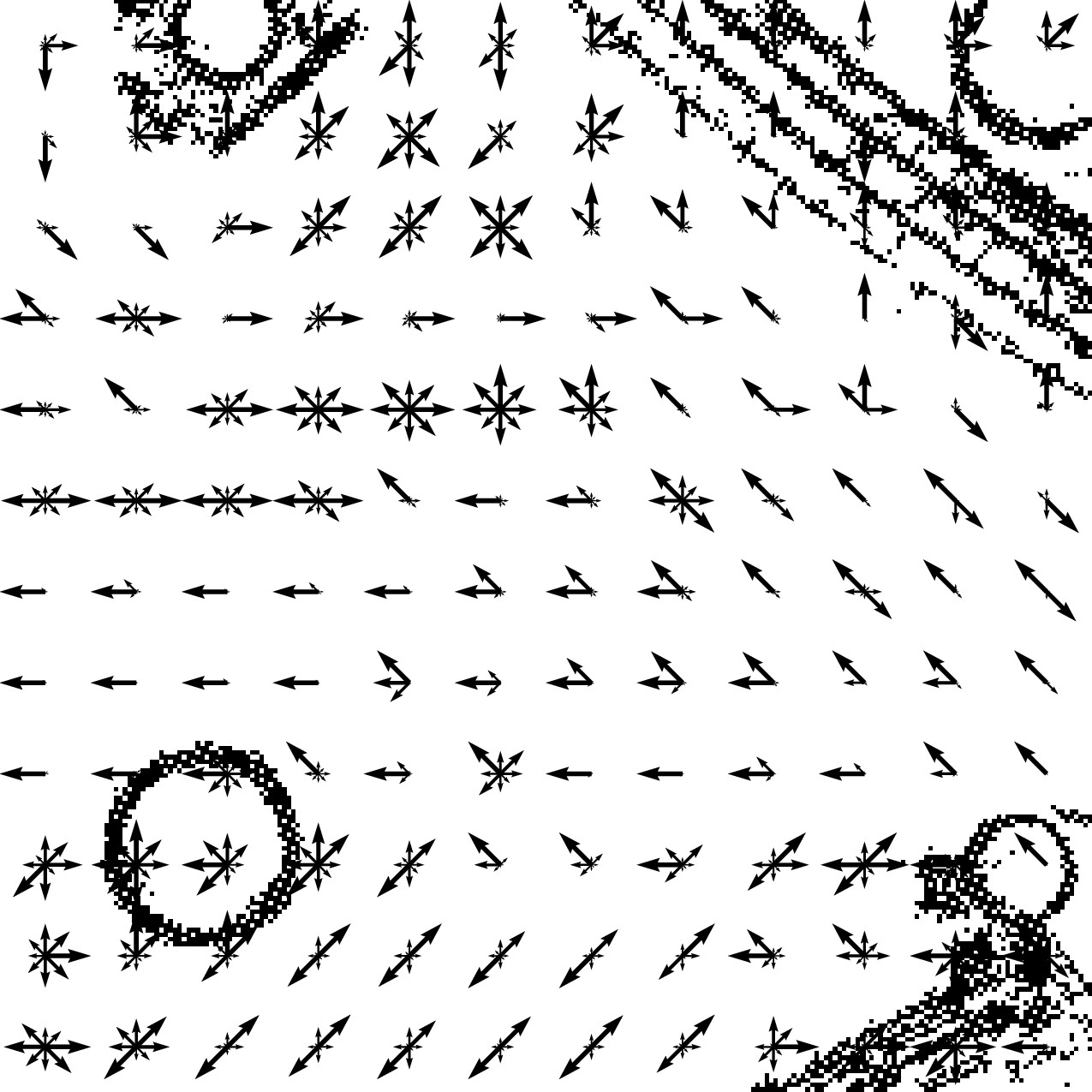}}
        \caption{ATC: BFF($\matr{\bar{d}}^W_{1.0}, \mathbb{D}^S[10000]$)}%
        \label{fig:quivers-atc-10k}
    \end{subfigure}
    \begin{subfigure}[t]{0.24\textwidth}
        \fbox{%
        \includegraphics[width=\linewidth]{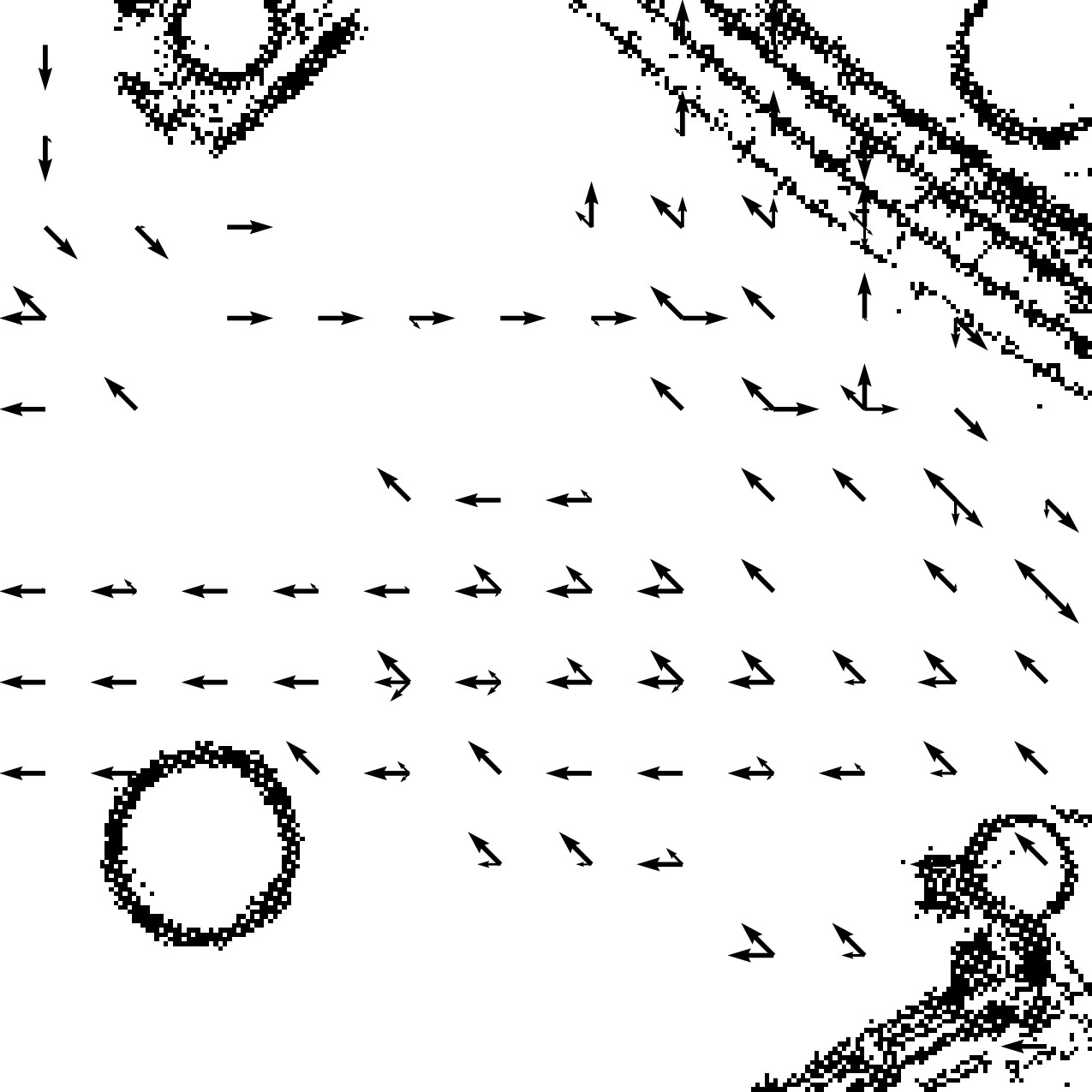}}
        \caption{ATC: FF($\mathbb{D}^S[10000]$)}%
        \label{fig:quivers-atc-noprior}
    \end{subfigure}
    \begin{subfigure}[t]{0.24\textwidth}
        \fbox{\includegraphics[width=\linewidth]{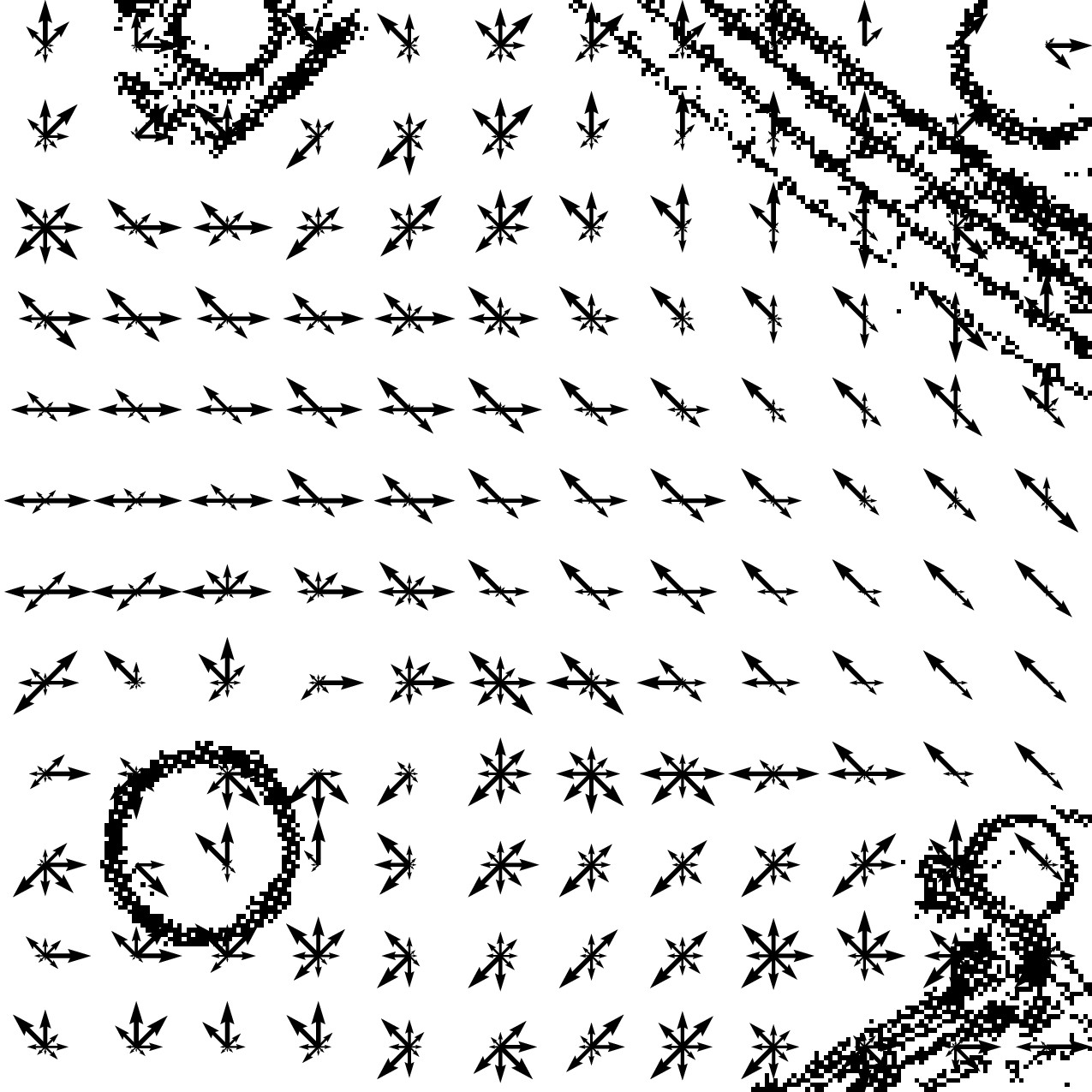}}
        \caption{ATC: Gold standard ($\ffv{}$)}\label{fig:quivers-atc-gt}
    \end{subfigure}\\[1em]
    \begin{subfigure}[t]{0.24\textwidth}
        \fbox{\includegraphics[width=\linewidth]{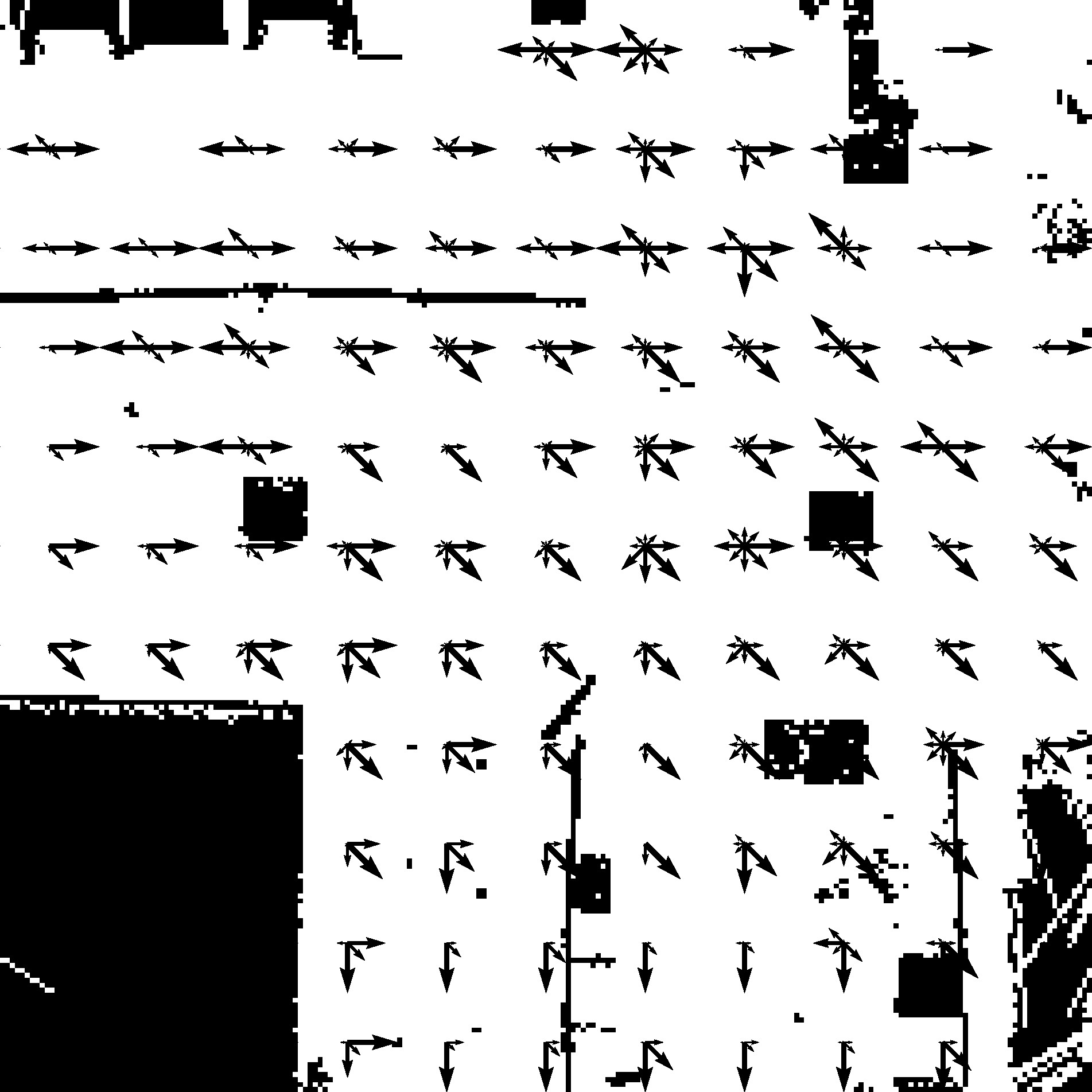}}
        \caption{KTH: Prior only ($\matr{\bar{d}}^W_{1.0}$)}%
        \label{fig:quivers-kth-prior}
    \end{subfigure}
    \begin{subfigure}[t]{0.24\textwidth}
        \fbox{\includegraphics[width=\linewidth]{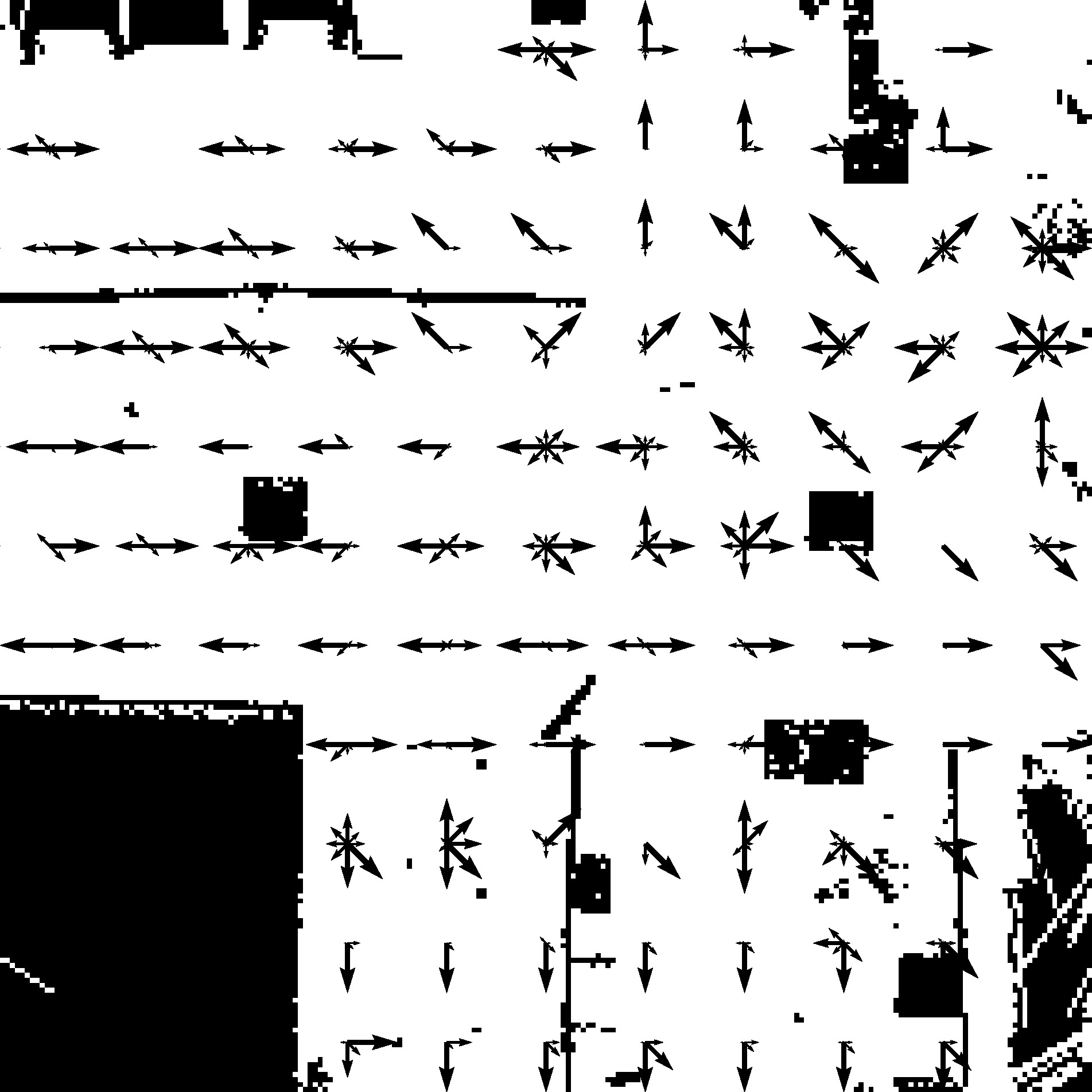}}
        \caption{KTH: BFF($\matr{\bar{d}}^W_{1.0}, \mathbb{D}^K[10000]$)}%
        \label{fig:quivers-kth-10k}
    \end{subfigure}
    \begin{subfigure}[t]{0.24\textwidth}
        \fbox{%
        \includegraphics[width=\linewidth]{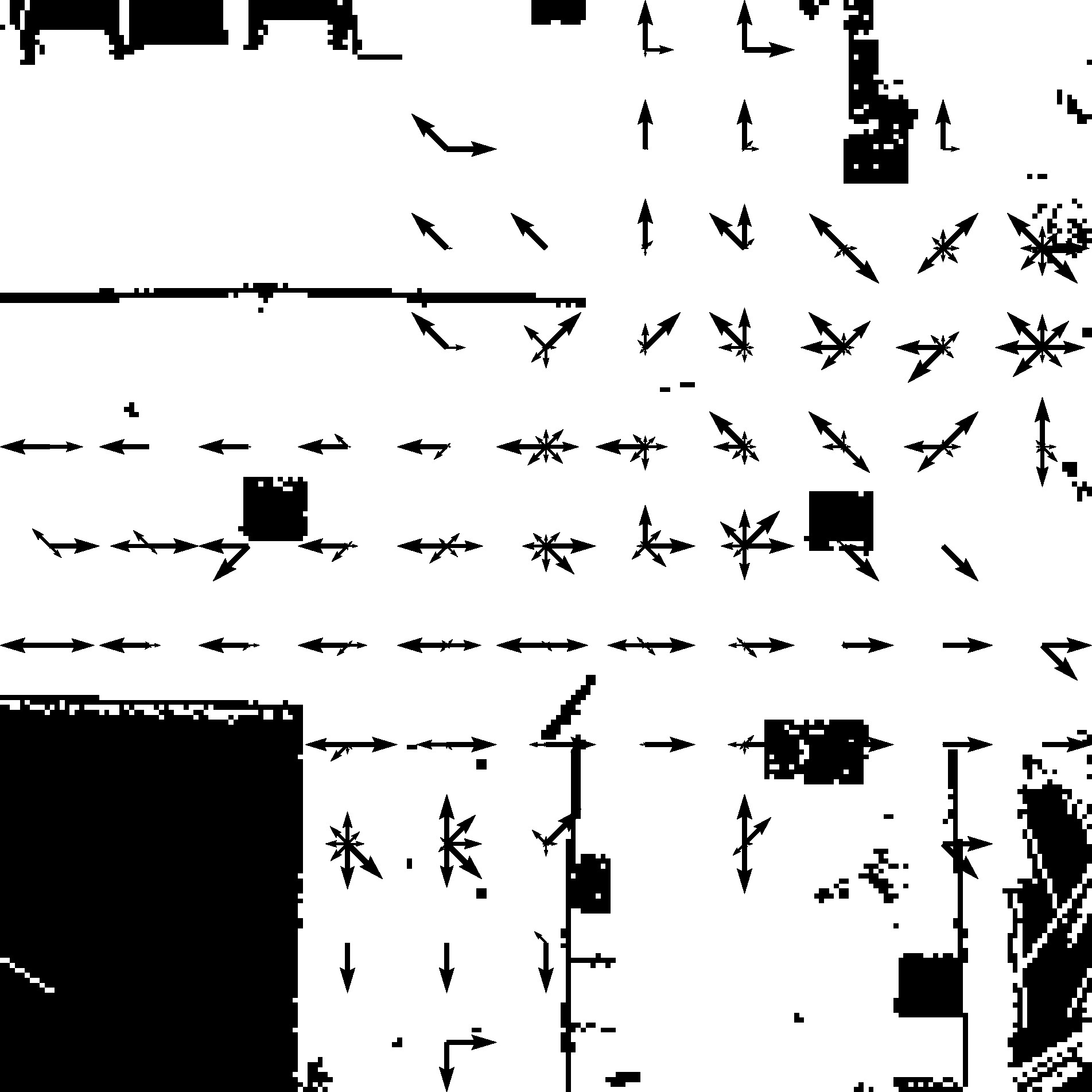}}
        \caption{KTH: FF($\mathbb{D}^K[10000]$)}%
        \label{fig:quivers-kth-noprior}
    \end{subfigure}
    \begin{subfigure}[t]{0.24\textwidth}
        \fbox{\includegraphics[width=\linewidth]{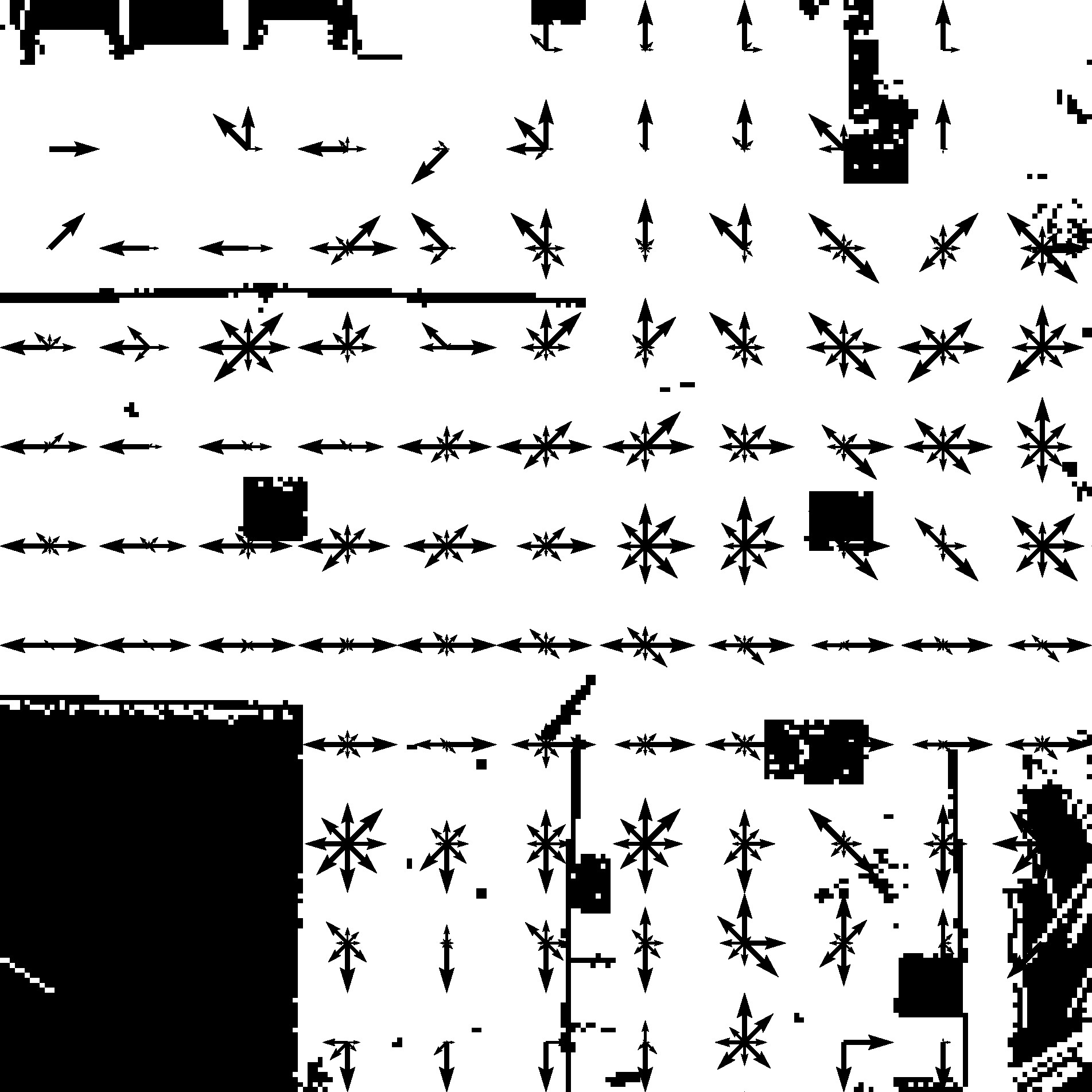}}
        \caption{KTH: Gold standard ($\ffk{}$)}\label{fig:quivers-kth-gt}
    \end{subfigure}
    \caption{Visualization of different \acp{mod} overlaid on the occupancy map.
    The first row represents a \qty[mode=text]{144}{\metre^2} area of ATC, while
    the second row represents a \qty[mode=text]{121}{\metre^2} area of KTH. From
    left to right, we show the model used as prior built at a resolution of
    \qty[mode=text]{1}{\metre\per cell} (a, e), the proposed Bayesian floor
    field model obtained by updating that prior with \num[mode=text]{10000}
    observations (b, f), the floor field model build with the same
    \num[mode=text]{10000} observations (c, g), and the gold standard floor
    field model built using all available observations in the each dataset (d,
    h).}\label{fig:quivers}
\end{figure*}

Finally, a visualization of different models is shown in \figref{fig:quivers},
from where few observations can be made. First, while the prior is clearly more
aligned to the gold standard on the different day on ATC
(\figref{fig:quivers-atc-prior}) than on KTH (\figref{fig:quivers-kth-prior}),
even on the latter it still captures some properties of the flow, \eg{} that a
portion of the flow will move into the opening at the bottom of the figure.
Moreover, when comparing the Bayesian model vs the floor field one
(\figref{fig:quivers-atc-10k} vs. \figref{fig:quivers-atc-noprior} and
\figref{fig:quivers-kth-10k} vs \figref{fig:quivers-kth-noprior}) it can be
noticed how both models fit the data similarly where data is abundant, but,
contrary to the floor field model, the Bayesian model is able to still provide
informed estimates where no observation is available by leveraging the prior. 

In conclusion, these results together show that occupancy-based maps of dynamics
can be effectively combined with trajectory data to learn better \acp{mod} with
less data. Considering that in both environments the floor field model starts
performing better than the proposed Bayesian model only after hundreds of people
have been observed, we see these results as extremely important for concrete
deployment of robots in new environments, where the expectation of availability
of hundreds of pre-recorded trajectories is unrealistic.

\section{Conclusions}
\label{sec:conclusions}

In this work, we presented a novel approach to infer \acfp{mod} from
architectural geometry and transfer the learned model to new unseen
environments. Moreover we proposed a mapping method using Bayesian inference to
combine occupancy-based and trajectory-based \acp{mod}. We evaluated the
generalization ability of the proposed method on human trajectories in different
large-scale environments, showing that, when tasked to predict trajectories
across environments, the proposed method is able to improve performance while
requiring less trajectories.

While the Bayesian approach presented here utilizes a prior that relies on the
environment occupancy, this is not a requirement. In the future, it will be
interesting to study if similar generalization can be obtained with other more
advanced methods to learn priors, \eg{} utilizing environmental semantic cues.

In conclusion, both the ability of the proposed method to generalize to unseen
large-scale buildings and its ability to combine different type of data is
unprecedented in \acp{mod} literature. When considering these findings from a
broader perspective, they illuminate an untapped potential in robotic mapping:
most mapping approaches tend to focus on one property of the environment at a
time, however modeling the latent correlation between different properties can
increase data efficiency and provide richer maps. Studying which environment
characteristics are good predictors for properties that are time-consuming or
expensive to map, like people flow, will be crucial to make complex maps more
ubiquitous in robotics.
\section*{Acknowledgements} 
\label{sec:acknowledge}

This work was supported by the Research Council of Finland, decision 354909.

We gratefully acknowledge the support of NVIDIA Corporation with the donation of
the Titan Xp GPU used for this research.


\newpage

\bibliographystyle{IEEEtran}
\bibliography{refs}

\end{document}